\documentclass[letterpaper]{article} 
\usepackage{aaai2026}  
\usepackage{times}  
\usepackage{helvet}  
\usepackage{courier}  
\usepackage[hyphens]{url}  
\usepackage{graphicx} 
\usepackage{natbib}  
\usepackage{amsmath, amssymb}
\usepackage{bm}
\usepackage{booktabs}
\usepackage{multirow} 
\usepackage{caption} 
\usepackage{algorithm}
\usepackage{algorithmic}

\usepackage{newfloat}
\usepackage{listings}
\DeclareCaptionStyle{ruled}{labelfont=normalfont,labelsep=colon,strut=off} 
\floatstyle{ruled}
\newfloat{listing}{tb}{lst}{}
\floatname{listing}{Listing}
\title{Learn from Global Correlations: Enhancing Evolutionary Algorithm\\ via Spectral GNN}
\author {
	Kaichen Ouyang\textsuperscript{\rm 1},
	Zong Ke\textsuperscript{\rm 2},
	Shengwei Fu\textsuperscript{\rm 3},
	Lingjie Liu \textsuperscript{\rm 4},
	Puning Zhao\textsuperscript{\rm 5},
	Dayu Hu\textsuperscript{\rm 6}
}
\affiliations {
	\textsuperscript{\rm 1}Department of Mathematics, University of Science and Technology of China, Hefei, China\\
	\textsuperscript{\rm 2}Faculty of Science, National University of Singapore, Singapore\\
	\textsuperscript{\rm 3}Key Laboratory of Advanced Manufacturing Technology, Ministry of Education, Guizhou University, Guiyang, China\\
	\textsuperscript{\rm 4}Mathematical Institute, University of Oxford, Oxford, United Kingdom\\
	\textsuperscript{\rm 5}School of Cyber Science and Technology, Sun Yat-sen University, Shenzhen, China\\
	\textsuperscript{\rm 6}College of Medicine and Biological Information Engineering, Northeastern University, Shenyang, China\\
	oykc@mail.ustc.edu.cn, hudy@bmie.neu.edu.cn
}

\usepackage{bibentry}

\begin{document}

\maketitle

\begin{abstract}
Evolutionary algorithms (EAs) are optimization algorithms that simulate natural selection and genetic mechanisms. Despite advancements, existing EAs have two main issues: (1) they rarely update next-generation individuals based on global correlations, thus limiting comprehensive learning; (2) it is challenging to balance exploration and exploitation, excessive exploitation leads to premature convergence to local optima, while excessive exploration results in an excessively slow search. Existing EAs heavily rely on manual parameter settings, inappropriate parameters might disrupt the exploration-exploitation balance, further impairing model performance.  To address these challenges, we propose a novel evolutionary algorithm framework called Graph Neural Evolution (GNE).   Unlike traditional EAs, GNE represents the population as a graph, where nodes correspond to individuals, and edges capture their relationships, thus effectively leveraging global information.   Meanwhile, GNE utilizes spectral graph neural networks (GNNs) to decompose evolutionary signals into their frequency components and designs a filtering function to fuse these components. High-frequency components capture diverse global information, while low-frequency components capture more consistent information. This explicit frequency filtering strategy directly controls global-scale features through frequency components, overcoming the limitations of manual parameter settings and making the exploration-exploitation control more interpretable and effective.  Extensive evaluations on nine benchmark functions (e.g., Sphere, Rastrigin, and Rosenbrock) demonstrate that GNE consistently outperforms both classical algorithms (GA, DE, CMA-ES) and advanced algorithms (SDAES, RL-SHADE) under various conditions, including original, noise-corrupted, and optimal solution deviation scenarios. GNE achieves solution quality several orders of magnitude better than other algorithms (e.g., 3.07e-20 mean on Sphere vs. 1.51e-07).
\end{abstract}

\begin{links}
    \link{Code}{https://github.com/oykc1234/GNE}
\end{links}

\section{Introduction}

Graph Neural Networks (GNNs) have been widely applied in fields such as social network analysis, recommendation systems, and protein structure prediction because of their capacity to effectively model complex relational structures using graph representations\cite{kipf2016semi,petar2018graph,xu2018powerful,liu2025phosf3c,lu2025transhla,zhou2025,cui2020surface}. In GNNs, dependencies between nodes are captured by propagating and aggregating information across the network\cite{10325611,10243081,SEC-LSRM,SAMVGC,MDRO}, thereby enabling adaptive learning and optimization based on node interactions. Similarly, biological evolution involves genetic interactions characterized by variation, recombination, and natural selection, which are processes that collectively facilitate the evolutionary optimization of genotypes toward greater fitness\cite{wright1932roles,darwin2023origin,brady1985optimization}. Given these parallels, we propose that GNNs can be understood through an evolutionary perspective, where the propagation of information among nodes mirrors the diversity-exploring mechanisms of crossover and mutation, as well as the adaptive optimization mechanism of directional selection in evolution.

Evolutionary adaptation thrives on two complementary strategies: exploration through genetic diversity (e.g., avian wing polymorphism spanning short, medium, and long morphotypes) and exploitation via trait optimization\cite{eiben1998evolutionary,vcrepinvsek2013exploration,yu2024balance}.Consider three avian archetypes: short-winged penguins excel in aquatic foraging, medium-winged gulls balance aerial maneuverability with diving efficiency, and long-winged albatrosses dominate transoceanic flight—each adaptation emerging from mutation-driven exploration and sustained through environmental selection. Similarly, giraffes’ neck elongation, refined iteratively over generations, exemplifies exploitation’s role in amplifying fitness-critical traits. Evolutionary algorithms (EAs) formalize this duality: crossover/mutation operators probe global solution spaces (exploration), while selection pressures refine local optima (exploitation)\cite{holland1992genetic,chao2024match,storn1997differential}.

Conventional EAs fail to fully exploit the global relationships among individuals in a population, relying on limited information from a few individuals for updating a single solution\cite{gorges1991genetic,vcrepinvsek2013exploration,opara2019differential}, thus hindering comprehensive learning. Additionally, existing EAs optimize functions in a black-box manner, with parameter selection relying on extensive experimental results. Inappropriate parameter choices can easily disrupt the balance between exploration and exploitation, potentially leading to suboptimal results \cite{ojha2022assessing,  tanabe2019reviewing,karafotias2014parameter}.To address these limitations, we reinterpret evolutionary algorithms through a GNN lens, proposing Graph Neural Evolution (GNE). In GNE, each candidate solution in  population is represented as a node in a graph, with the corresponding value in the adjacency matrix between two nodes reflecting the similarity of them. This approach fully leverages global correlations among individuals within the population, overcoming the constraints of local updates. These global correlations resemble biological reproductive isolation, where genetically similar individuals preferentially exchange traits\cite{dobzhansky1982genetics}. Moreover, we compute the normalized Laplacian matrix from the population's adjacency matrix and decompose it into frequency components through its eigenvalues and eigenvectors in the spectral domain. Similar to spectral GNNs, GNE balances high-frequency exploration (diverse traits) and low-frequency exploitation (consistent traits) via filter functions. Notably, this explicit spectral aggregation strategy can control global-scale features through frequency components, making the exploration–exploitation control both interpretable and effective.

This spectral perspective provides a unified mathematical framework: the balance between exploration and exploitation in evolutionary algorithms is equivalent to designing an appropriate filter that assigns suitable weights to the different frequency components of the population, where strengthening the high-frequency components corresponds to reinforcing exploration behavior, and strengthening the low-frequency components corresponds to reinforcing exploitation behavior. By formalizing this duality, GNE provides interpretable control over the exploration-exploitation balance while benefiting from global correlations. Through rigorous evaluations on nine benchmark functions (e.g., Sphere, Rastrigin, and Rosenbrock), GNE consistently outperforms classical algorithms, including Genetic Algorithm (GA) \cite{holland1992genetic}, Differential Evolution (DE) \cite{storn1997differential}, Covariance Matrix Adaptation Evolution Strategy (CMA-ES) \cite{hansen2003reducing}, and advanced algorithms such as Evolution Strategy Based on Search Direction Adaptation (SDAES) \cite{he2019large} and Learning Adaptive Differential Evolution by Natural Evolution Strategies (RL-SHADE) \cite{zhang2022learning} across scenarios with original, noise-corrupted, and deviated-optimum conditions. GNE leads other algorithms by multiple orders of magnitude in solution quality (e.g., 3.07e-20 vs. 1.51e-07 mean on Sphere). This framework is broadly applicable to complex optimization problems in the real world, such as engineering design, and optimization of parameters and architecture of neural networks\cite{cai2019efficient,zhou2021survey,li2023survey}.To summarize, our contributions are threefold:

\textbf{Problem.} We discuss the intrinsic duality between Spectral GNNs and EAs. To the best of our knowledge, this is the first comprehensive study focusing on graph neural evolution.

\textbf{Algorithm.} Different from existing EAs, we construct an adjacency matrix to capture global correlations within the population. The global evolutionary signals are decomposed into their frequency components in the spectral domain and then fused via bespoke filter functions. This explicit spectral aggregation strategy can control global features through frequency components, making the exploration-exploitation control both interpretable and effective.

\textbf{Evaluation.} Extensive empirical studies have been conducted on nine representative benchmark functions to evaluate the performance of GNE. The experimental results demonstrate that it can effectively learn from global correlations and balance exploration and exploitation, thereby significantly outperforming existing EAs in terms of accuracy, convergence, stability, and robustness.

\section{Related Works}

The proposed GNE framework integrates EAs with the GNNs by modeling population dynamics as spectral graph operations. We first review related works on EAs and GNNs, identifying gaps that motivate GNE.

\begin{figure*}[t]
	\centering
	\includegraphics[width=0.8\textwidth]{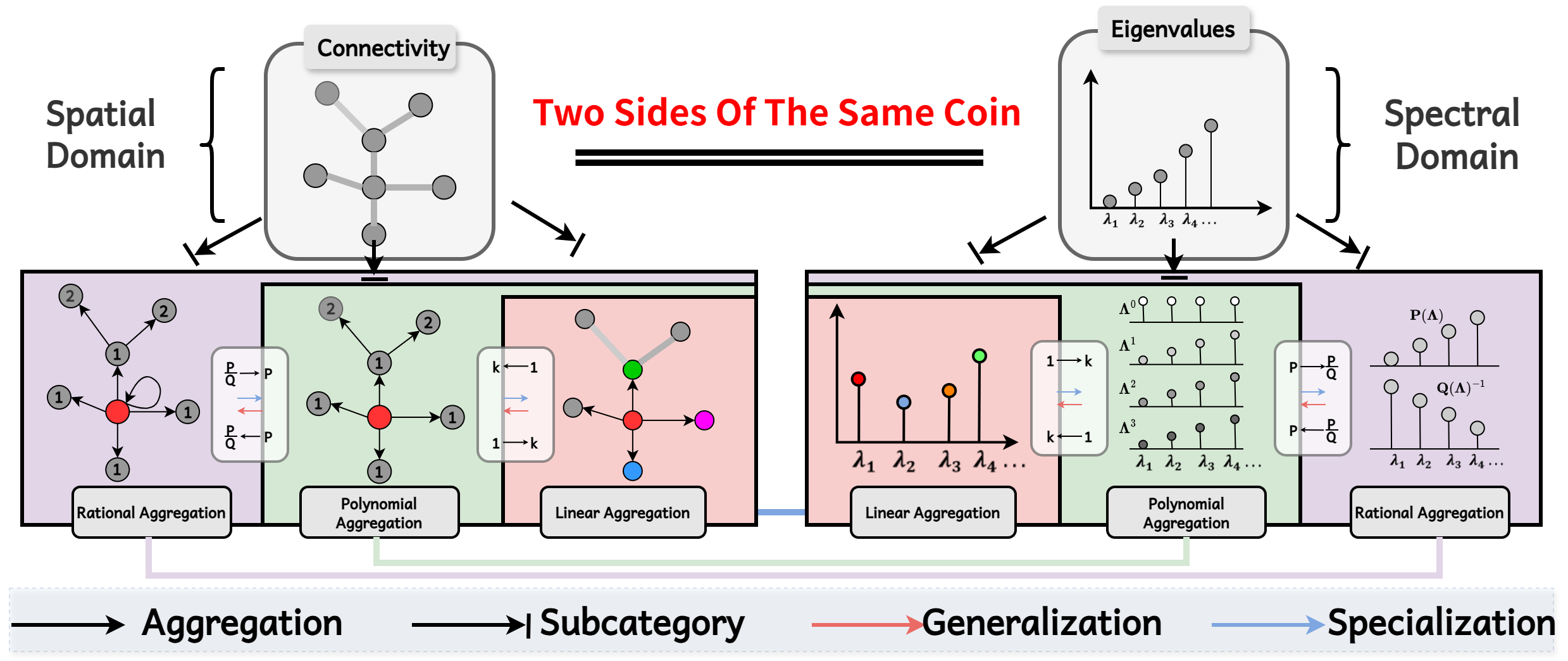}
	\caption{The relationship between spatial and spectral domain graph neural networks.}
	\vspace{-10pt}    
	\label{spavsspec}
\end{figure*}

\textbf{Evolutionary Algorithms.} EAs are a family of metaheuristic optimization techniques inspired by natural evolutionary processes, incorporating operators such as natural selection and mutation\cite{ouyang2025beaver,ouyang2025dynamic}. GA is one of the earliest and most representative EAs, and is particularly effective in solving discrete and multimodal optimization problems\cite{holland1992genetic}. In recent years, DE\cite{storn1997differential,opara2019differential,pant2020differential} and Evolution Strategy (ES)\cite{beyer2002evolution,lange2024large,beyer2001theory} have emerged as prominent research domains. DE generates candidate solutions through differential mutation and selects the fittest among them. Numerous variants have been developed, including Adaptive Differential Evolution with Optional External Archive (JADE) \cite{zhang2009jade}, which incorporates external archives and adaptive parameter updates; Success-History Based Parameter Adaptation for Differential Evolution (SHADE) \cite{tanabe2013success}, which utilizes historically successful parameter settings to refine control parameters; and RL-SHADE \cite{zhang2022learning}, which further enhances DE performance by integrating natural evolution strategies for more effective parameter optimization. Among ES methods, Covariance Matrix Adaptation Evolution Strategy (CMA-ES) is one of the most widely recognized algorithms \cite{hansen2003reducing}, adjusting the covariance matrix to effectively explore the solution space. Several variants have been developed, including OpenES for large-scale parallel computation \cite{salimans2017evolution}; Parameter-Exploring Policy Gradients (PEPG) for estimating gradients in the parameter space \cite{sehnke2010parameter};Mixture Model based Evolution Strategy (MMES) \cite{he2020mmes}, which accelerates CMA-ES in large-scale optimization tasks; and SDAES \cite{he2019large}, which reduces both time and space complexity in high-dimensional settings. From GA to DE, ES, and their variants, the design of evolutionary algorithms has consistently centered on balancing exploration and exploitation. Exploration introduces diversity to expand the search space and prevent premature convergence to local optima, while exploitation refines promising solutions to gradually approach the global optimum\cite{eiben1998evolutionary}.

\textbf{Graph Neural Networks.} GNNs process graph-structured data from two principal perspectives: spatial and spectral\cite{chen2023bridging,muhammet2020spectral,wang2022powerful,meng2024fedean,meng2024sarf,yang2021interpretable,yang2022self}. Spatial GNNs are more widely known, with the Graph Convolutional Network (GCN) \cite{kipf2016semi} being a representative example, which aggregates information from neighboring nodes to learn local structures. The Graph Attention Network (GAT) \cite{petar2018graph} further enhances GCNs by incorporating self-attention mechanisms, while GraphSAGE generates node embeddings by sampling and aggregating features from local neighborhoods, thereby enabling generalization to previously unseen nodes \cite{hamilton2017inductive}. In contrast, spectral GNNs rely on the graph Laplacian and Fourier transform for convolution operations, thus learning from global structural patterns. Here we review representative spectral methods. ChebyNet introduces Chebyshev polynomials to construct efficient spectral filters, thereby improving scalability \cite{defferrard2016convolutional}. BernNet learns spectral filters through Bernstein polynomial approximations, which provide flexibility in filter design and help overcome the limitations of predefined or unconstrained filters \cite{he2021bernnet}. JacobiConv increases expressive power by employing an orthogonal Jacobi basis, yielding superior performance over other spectral GNNs across both synthetic and real-world datasets \cite{wang2022powerful}. In brief, spectral GNNs derive filtering expressions that modulate both the low-frequency and high-frequency components of the original graph signals. This process aligns with evolutionary dynamics: low-frequency components represent the stable core of the graph signal, while high-frequency components capture variations among nodes, thereby capturing finer details.

\textbf{Bridging the Gap.} Prior efforts rarely unify EAs and GNNs. Conventional EAs are black-box methods and ignore graph-structured interactions, while GNNs focus on representation learning rather than evolutionary optimization. Our work addresses this gap by reinterpreting EA dynamics as spectral graph filtering, enabling explicit control over exploration–exploitation through frequency modulation using filter functions.
\begin{figure*}[t]
	\centering
	\includegraphics[width=0.7\textwidth]{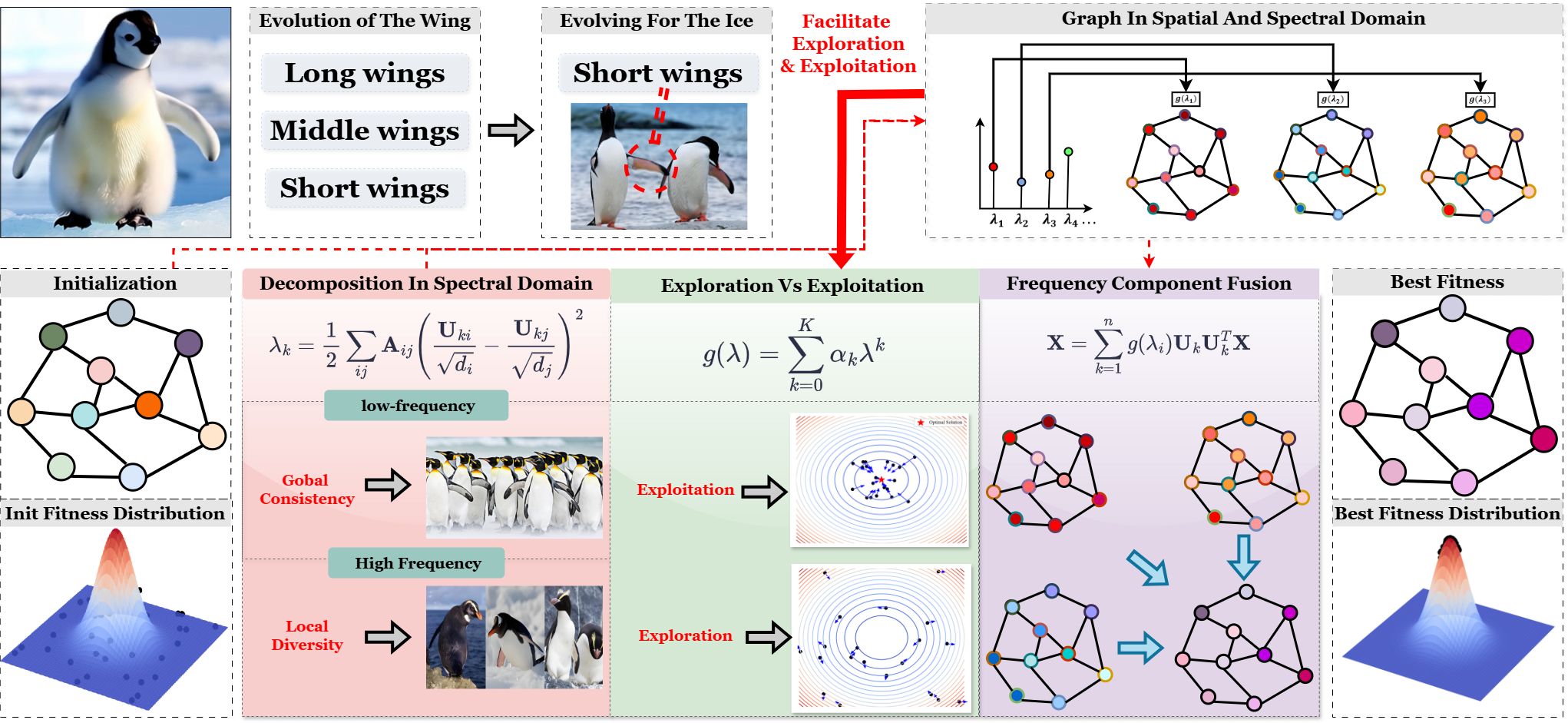}
	\caption{The framework of the proposed GNE method. It begins by constructing the population's adjacency matrix to capture global relationships, which is computed using cosine similarities between individual nodes.The obtained adjacency matrix is further transformed into the graph Laplacian. Through this process, GNE extracts frequency components that characterize population diversity (high-frequency) and consistency (low-frequency). Furthermore, exploration–exploitation balance is achieved by modulating the weights of these frequency components with a polynomial filter, while a sampling function resamples the population in feature space to optimize overall fitness.}
	\vspace{-10pt}
	\label{fig_fra}
\end{figure*}

\section{Evolution via Spectral Graph Representations}
\subsection{Preliminary}
\textbf{Notations.} Given a set of nodes $\mathcal{V}=\{v_1,\dots,v_N\}$, an undirected graph can be represented in matrix form as $\mathcal{G}=(\mathbf{X},\mathbf{A})$, where the adjacency matrix $\mathbf{A}\in\mathbb{R}^{N\times N}$ encodes the pairwise interactions between nodes.The degree matrix $\mathbf{D} \in \mathbb{R}^{N \times N}$ is defined as $\mathbf{D}_{ii} = \sum_j \mathbf{A}_{ij}
$, and the normalized Laplacian matrix is presented as $\widehat{\mathbf{L}}\ = \mathbf{I} - \mathbf{D}^{-\frac{1}{2}} \mathbf{A} \mathbf{D}^{-\frac{1}{2}}$. This matrix can also be expanded in the spectral form as $\widehat{\mathbf{L}}\ = \mathbf{U} \mathbf{\Lambda} \mathbf{U}^\top$, where $\mathbf{U}$ contains orthonormal eigenvectors and $\mathbf{\Lambda} = \mathrm{diag}(\lambda_1, \dots, \lambda_N)$ is a diagonal matrix of the corresponding eigenvalues, which satisfy $\lambda_k \in [0, 2]$. The graph Fourier transform is defined as $\widetilde{\mathbf{X}} = \mathbf{U}^\top \mathbf{X}$, with inverse transform $\mathbf{X} = \mathbf{U} \widetilde{\mathbf{X}}$. The component at frequency $\lambda_k$ is given by $\widetilde{\mathbf{X}}_{\lambda_k} = \mathbf{U}_k\mathbf{U}_k^\top \mathbf{X}
$, where $\mathbf{U}_k$ is the eigenvector corresponding to $\lambda_k$.
The spectral filter is defined as $g(\mathbf{\Lambda}) = \mathrm{diag}\bigl(g(\lambda_1), \dots, g(\lambda_N)\bigr)$ and is generally parameterized by a $k$th-order polynomial, $g(\lambda) = \sum_{k=0}^K \alpha_k \lambda^k$, which lifts to the vertex domain through $g(\widehat{\mathbf{L}}) = \mathbf{U} g(\mathbf{\Lambda}) \mathbf{U}^\top$.

\noindent\textbf{Problem formulation.} In the proposed GNE framework, the evolution of the population can be regarded as a frequency filtering process. By controlling the frequency components, the balance between exploration and exploitation is achieved. Specifically, the GNE population-update procedure is based on a polynomial filter and can be defined as follows:
\begin{equation}
	\begin{split}
		g(\widehat{\mathbf{L}})\,\mathbf{X} = \mathbf{U}\,g(\mathbf{\Lambda})\,\mathbf{U}^{\top}\mathbf{X} = \\
		\sum_{k=1}^n g(\lambda_k)\,\mathbf{U}_k\mathbf{U}_k^{\top}\mathbf{X} = \sum_{k=1}^n g(\lambda_k)\,\widetilde{\mathbf{X}}_{\lambda_k}
	\end{split}
\end{equation}
Here, $\mathbf{X}$ is integrated with the filtering function $g(\cdot)$ to produce the next generation of the population. This carefully designed population-update method introduces frequency regulation into the evolutionary framework, thereby rendering the entire update process both interpretable and effective.

\subsection{The proposed GNE model}
Existing representative evolutionary algorithms, such as DE\cite{storn1997differential}, typically operate via three key steps: mutation, crossover, and selection. During the mutation step, a mutant vector \(\bm{v}_i\) is generated for each individual \(\bm{x}_i\) in the population:
\begin{equation}
	\bm{v}_i = \bm{x}_{r1} + \alpha_s \cdot (\bm{x}_{r2} - \bm{x}_{r3})
\end{equation}
where \( \bm{x}_{r1} \), \( \bm{x}_{r2} \), and \( \bm{x}_{r3} \) denote distinct individuals randomly selected, and \( \alpha_s \) serves as the scaling factor that controls the magnitude of exploration. Then the crossover step generates a trial vector \( \bm{u}_i \) by merging components from both the mutant vector \( \bm{v}_i \) and the target vector \( \bm{x}_i \): 
\begin{equation}
	\bm{u}_{i,j} = \begin{cases} 
		\bm{v}_{i,j} & \text{if } rand(0,1) \leq \text{p}_{cr} \text{ or } j = j_{r} ~, \\
		\bm{x}_{i,j} & \text{otherwise}
	\end{cases}
\end{equation}
where \( \text{p}_{cr} \) is the crossover rate determining the probability of inheriting components from the mutant vector,  and \( j_{r} \) ensures that at least one component is inherited from \( \bm{v}_i \). Besides, \( \bm{u}_{i,j} \),
\( \bm{v}_{i,j} \) and \( \bm{x}_{i,j} \) denote the j-th components of the trial vector \(\bm{u}_i \), mutant vector \(\bm{v}_i \) and target vector $\bm{x}_i$, respectively. The selection step retains the superior individual between the trial vector \( \bm{u}_i \) and the target vector \( \bm{x}_i \):
\begin{equation}
	\mathbf{x}_i' = \begin{cases} 
		\mathbf{u}_i & \text{if } f(\mathbf{u}_i) \leq f(\mathbf{x}_i)~, \\
		\mathbf{x}_i & \text{otherwise}
	\end{cases}
\end{equation}
where \( \mathbf{x}_i' \) represents the updated individual for the next generation, and \( f(\cdot) \) denotes the fitness function. Higher values of \( \alpha_s \) and \( \text{p}_{cr} \) promote exploration, whereas lower values favor exploitation. However, these operations occur at the individual level, and for specific problem types, the algorithm's performance is sensitive to the choice of \( \alpha_s \) and \( \text{p}_{cr} \). While traditional EAs, such as DE, perform well on many problems, they lack modeling of the global correlations of the population, providing limited direct control and interpretability over the exploration-exploitation trade-off. These limitations may result in instability across different problem types.

To address these limitations, we propose the GNE, which reformulates the EAs within a graph-based framework to capture global correlations (Figure \ref{fig_fra}). In GNE, individuals are represented as nodes, and their interactions are modeled using an adjacency matrix. The adjacency matrix \( \mathbf{A} \) captures the strength of node interactions, prioritizing exchanges among similar individuals, analogous to reproductive isolation mechanisms in biology. This matrix is computed using the cosine similarity between individuals \( i \) and \( j \), based on their difference vectors \( \mathbf{z}_i = \bm{x}_i - \bm{x}_0 \), where \( \bm{x}_0 \) denotes the population centroid. The element \(\mathbf{A}_{ij}\) represents the \(i\)-th row and \(j\)-th column of the adjacency matrix:
\begin{equation}
	\mathbf{A}_{ij} = \frac{\mathbf{Z}_{i:}\mathbf{Z}_{j:}^\top}{\|\mathbf{Z}_{i:}\|\|\mathbf{Z}_{j:}\|}
\end{equation}
\begin{figure*}[t]
	\centering
	\includegraphics[width=0.8\textwidth]{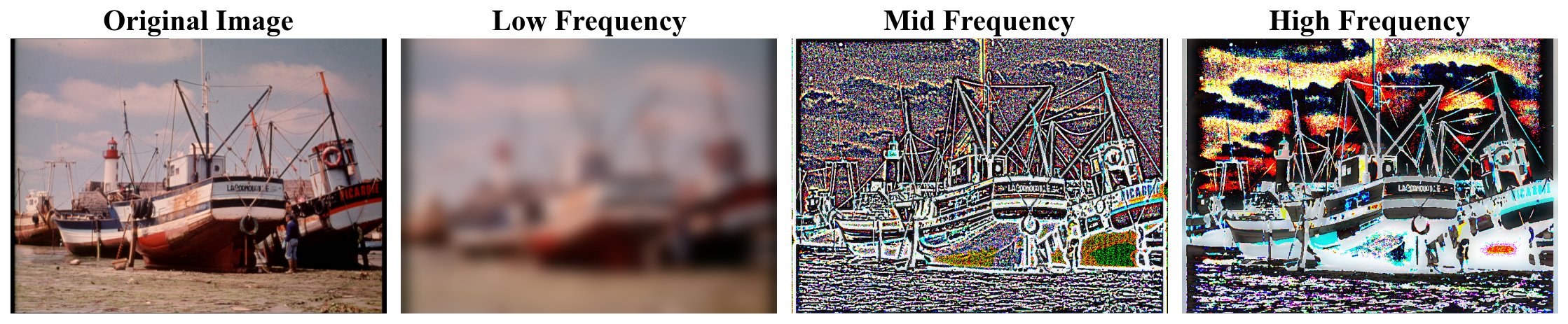}
	\caption{Visualizing the relationship between consistent traits (low-frequency) and diverse traits (high-frequency) in the Boat.}
	\vspace{-10pt}
	\label{fig_vis}
\end{figure*}
Based on the obtained adjacency matrix, we further compute the normalized graph Laplacian matrix \( \widehat{\mathbf{L}} = \mathbf{I}-\mathbf{D}^{-\frac{1}{2}}\mathbf{A}\mathbf{D}^{-\frac{1}{2}} \), which undergoes spectral decomposition \( \widehat{\mathbf{L}} = \mathbf{U}\mathbf{\Lambda}\mathbf{U}^\top \) to generate eigenvalues \( \lambda_k \), which are given by:
\begin{equation}
	\lambda_k = \frac{1}{2} \sum_{ij} \mathbf{A}_{ij} \left( \frac{\mathbf{U}_{ki}}{\sqrt{d_i}} - \frac{\mathbf{U}_{kj}}{\sqrt{d_j}} \right)^2
\end{equation}
\begin{algorithm}[t]
	\caption{Iterative Algorithm of GNE}
	\label{alg:gne}
	\begin{algorithmic}[1]
		\STATE \textbf{Input:} Population size $N$, maximum number of iterations $T$, sampling functions $\varphi(\cdot)$ and $\phi(\cdot)$, spectral filter $g(\cdot)$, objective function $f(\cdot)$.
		\STATE \textbf{Output:} Optimal solution $\mathbf{x}_{\mathrm{best}}$, objective value $f_{\mathrm{best}}$.
		\STATE Initialize population $\mathbf{X} = \{\mathbf{x}_1, \dots, \mathbf{x}_N\}$ randomly.
		\STATE Evaluate $f(\mathbf{x}_i)$ for each $\mathbf{x}_i \in \mathbf{X}$.
		\STATE $\mathbf{x}_{\mathrm{best}} \leftarrow \arg\min_{\mathbf{x}_i \in \mathbf{X}} f(\mathbf{x}_i)$.
		\STATE $f_{\mathrm{best}} \leftarrow f(\mathbf{x}_{\mathrm{best}})$.
		\FOR{$t = 1$ \textbf{to} $T$}
		\STATE Encode population with $\psi(\cdot)$: $\mathbf{X} \leftarrow \psi(\cdot)(\mathbf{X})$.
		\STATE Compute adjacency matrix $\mathbf{A}$ from $\mathbf{X}$.
		\STATE Compute degree matrix $\mathbf{D}$ of $\mathbf{A}$.
		\STATE Compute normalized $\widehat{\mathbf{L}}=\mathbf{I} - \mathbf{D}^{-1/2} \mathbf{A} \mathbf{D}^{-1/2}$.
		\STATE Compute eigen-decomposition $\widehat{\mathbf{L}} = \mathbf{U}\mathbf{\Lambda}\mathbf{U}^\top$.
		\STATE Update population using spectral filter: $\hat{\mathbf{X}} \leftarrow \mathbf{U}\,g(\mathbf{\Lambda})\,\mathbf{U}^\top \mathbf{X}$.
		\STATE Apply encoder $\phi$ to $\mathbf{X}$: $\mathbf{X} \leftarrow \phi(\hat{\mathbf{X}})$.
		\STATE Evaluate $f(\mathbf{x}_i)$ for each $\mathbf{x}_i \in \mathbf{X}$.
		\STATE Set $\mathbf{x}' \leftarrow \arg\min_{\mathbf{x}_i \in \mathbf{X}} f(\mathbf{x}_i)$.
		\IF{$f(\mathbf{x}') < f_{\mathrm{best}}$}
		\STATE $\mathbf{x}_{\mathrm{best}} \leftarrow \mathbf{x}'$.
		\STATE $f_{\mathrm{best}} \leftarrow f(\mathbf{x}_{\mathrm{best}})$.
		\ENDIF
		\ENDFOR
		\STATE \textbf{Return:} $\mathbf{x}_{\mathrm{best}}$, $f_{\mathrm{best}}$.
	\end{algorithmic}
\end{algorithm}
where \( \mathbf{U}_{ki} \) and \( \mathbf{U}_{kj} \) denote the components of individuals \( i \) and \( j \) in the eigenvector \( \mathbf{U}_k \), and \( d_i \) and \( d_j \) represent their degrees. The eigenvalues \( \lambda_k \) have a physical meaning as frequencies and quantify the variability along \( \mathbf{U}_k \). Higher values of \( \lambda_k \) indicate greater variability, with transitions from high to low frequencies reflecting a shift from diverse traits to consistent traits, which aligns with maintaining population diversity in EAs to fully explore the solution space while preserving consistency for fine exploration around the optimal solution. Figure \ref{fig_vis} provides a vivid example visualizing this relationship.

By leveraging global correlations encoded in the eigenvalues $\lambda_k$, GNE directly regulates the exploration–exploitation trade-off within the population through the polynomial filter function $g(\lambda)=\sum_{k=0}^{K}\alpha_{k}\lambda^{k}$, as expressed by the following equation:
\begin{equation}
	\hat{\mathbf{X}} = \mathbf{U} g(\mathbf{\Lambda}) \mathbf{U}^\top \mathbf{X} =\sum_{k=1}^n g(\lambda_k)\,\mathbf{U}_k\mathbf{U}_k^{\top}\mathbf{X}=\sum_{k=1}^n g(\lambda_k)\,\widetilde{\mathbf{X}}_{\lambda_k}
\end{equation}
where the term \( \widetilde{\mathbf{X}}_{\lambda_k} \) represents the population's component corresponding to \( \lambda_k \), while \( g(\lambda_k) \) is the weight applied to this component. Increasing the weight of high-frequency components boosts exploration, while increasing the weight of low-frequency components strengthens exploitation. The updated population \( \hat{\mathbf{X}} \) is obtained by integrating different frequency components.
GNNs typically employ multilayer perceptron (MLP) encoders $\psi(\cdot)$ and $\phi(\cdot)$ to transform and map features both before and after filtering, thereby further optimizing node representations. Similarly, GNE treats $\psi(\cdot)$ and $\phi(\cdot)$ as sampling functions that guide the population to generate new individuals toward optimizing overall population fitness. Here, $\psi(\cdot)$ is an identity transformation, while $\phi(\cdot)$ is a Gaussian distribution centered around the elite individuals of the population, with a standard deviation $\sigma_t$ that adaptively varies with the iteration count $t$.To summarize, the overall update formula for GNE can be expressed as follows:
\begin{equation}
	\mathbf{X} \leftarrow \phi(\mathbf{U} g(\mathbf{\Lambda}) \mathbf{U}^{\top} \psi(\mathbf{X}) )
\end{equation}
To provide an intuitive overview of GNE’s update procedure, Algorithm \ref{alg:gne} illustrates the iterative update process.
\begin{table*}[t]
	\renewcommand\arraystretch{1.2}
	\caption{Performance Comparison of GNE with Other Evolutionary Algorithms on Benchmark Functions}
	\label{tab_1}
	\centering
	\resizebox{0.9\textwidth}{!}{
		\begin{tabular}{c|lcccccc}
			\toprule[1pt]
			\specialrule{0em}{0.5pt}{0.5pt}
			\toprule[1pt]
			\multirow{2}{*}{\textbf{Noise}} 
			& \multirow{2}{*}{\textbf{Function}}
			& \multicolumn{5}{c}{\textbf{Non-Graph EAs}}
			& \multicolumn{1}{c}{\textbf{Graph EAs}} \\
			\cmidrule(lr){3-7} \cmidrule(lr){8-8}
			& 
			& GA & DE & CMA-ES & SDAES & RL-SHADE & GNE \\
			\midrule[1pt]
			\multirow{9}{*}{\rotatebox{90}{\textbf{Without Noise}}}
			& Sphere        & 2.31E+04$_{\pm7.10E+03}$ & 4.53E-04$_{\pm1.75E-04}$ & 1.22E-05$_{\pm9.27E-06}$ & 9.22E-18$_{\pm1.43E-17}$ & 1.51E-07$_{\pm2.17E-07}$ & \textbf{3.07E-20$_{\pm5.26E-21}$} \\
			& Schwefel      & 5.60E+01$_{\pm8.58E+00}$ & 2.31E-03$_{\pm5.83E-04}$ & 5.28E-03$_{\pm1.46E-03}$ & 1.54E-06$_{\pm5.94E-06}$ & 4.14E-04$_{\pm4.99E-04}$ & \textbf{7.65E-10$_{\pm4.08E-11}$} \\
			& Schwefel 2.22 & 5.42E+04$_{\pm1.38E+04}$ & 3.24E+04$_{\pm5.31E+03}$ & 4.28E-01$_{\pm2.33E-01}$ & 1.35E-01$_{\pm9.86E-02}$ & 1.85E+01$_{\pm1.51E+01}$ & \textbf{6.40E-20$_{\pm1.33E-20}$} \\
			& Schwefel 2.26 & 7.22E+01$_{\pm7.57E+00}$ & 1.25E+01$_{\pm1.56E+00}$ & 2.35E-02$_{\pm6.63E-03}$ & 2.67E-03$_{\pm8.72E-03}$ & 5.03E+00$_{\pm2.29E+00}$ & \textbf{6.86E-11$_{\pm6.04E-12}$} \\
			& Rosenbrock    & 2.53E+07$_{\pm1.57E+07}$ & 1.51E+02$_{\pm5.00E+01}$ & 1.73E+02$_{\pm4.19E+02}$ & 4.90E+01$_{\pm5.87E+01}$ & 4.60E+01$_{\pm4.01E+01}$ & \textbf{2.85E+01$_{\pm4.19E-02}$} \\
			& Quartic       & 1.16E+01$_{\pm7.72E+00}$ & 5.84E-02$_{\pm1.35E-02}$ & 2.64E-02$_{\pm5.59E-03}$ & 1.31E-01$_{\pm5.25E-02}$ & 3.74E-02$_{\pm1.92E-02}$ & \textbf{7.24E-05$_{\pm5.65E-05}$} \\
			& Rastrigin     & 1.97E+01$_{\pm4.50E+01}$ & 5.69E-03$_{\pm7.74E+00}$ & 1.09E-03$_{\pm7.24E+01}$ & 5.85E+00$_{\pm3.22E+01}$ & 2.28E+00$_{\pm3.99E+00}$ & \textbf{1.28E-10$_{\pm0.00E+00}$} \\
			& Ackley        & 2.06E+02$_{\pm5.55E-01}$ & 7.85E-03$_{\pm1.17E-03}$ & 1.72E-04$_{\pm2.51E-04}$ & 1.07E-03$_{\pm9.09E+00}$ & 1.00E-02$_{\pm6.03E-01}$ & \textbf{0.00E+00$_{\pm1.01E-11}$} \\
			& Levy          & 1.78E+07$_{\pm6.49E+01}$ & 5.83E-05$_{\pm1.08E-02}$ & 1.29E-06$_{\pm7.96E-05}$ & 3.11E-02$_{\pm3.40E-03}$ & 2.97E-01$_{\pm1.03E-02}$ & \textbf{2.23E-03$_{\pm0.00E+00}$} \\
			\midrule[1pt]
			\multirow{9}{*}{\rotatebox{90}{\textbf{With Noise}}}
			& Sphere    & 2.31E+04$_{\pm7.44E+03}$ & 2.87E-01$_{\pm4.21E-02}$ & 1.12E-01$_{\pm2.67E-02}$ & 3.64E-01$_{\pm9.54E-02}$ & 2.13E-01$_{\pm7.92E-02}$ & \textbf{1.11E-04$_{\pm8.19E-05}$} \\
			& Schwefel  & 5.49E+01$_{\pm8.94E+00}$ & 6.43E-01$_{\pm7.48E-02}$ & 3.78E-01$_{\pm4.18E-02}$ & 2.51E+00$_{\pm2.29E+00}$ & 6.04E-01$_{\pm2.95E-01}$ & \textbf{1.96E-04$_{\pm1.84E-04}$} \\
			& Schwefel 2.22 & 5.52E+04$_{\pm1.66E+04}$ & 3.35E+04$_{\pm5.03E+03}$ & 6.30E-01$_{\pm2.22E-01}$ & 8.72E+00$_{\pm3.38E+00}$ & 1.76E+01$_{\pm1.16E+01}$ & \textbf{9.12E-05$_{\pm7.37E-05}$} \\
			& Schwefel 2.26 & 7.27E+01$_{\pm9.53E+00}$ & 1.58E+01$_{\pm2.94E+00}$ & 4.10E-01$_{\pm5.56E-02}$ & 4.54E+00$_{\pm2.39E+00}$ & 1.04E+01$_{\pm3.62E+00}$ & \textbf{1.56E-04$_{\pm1.15E-04}$} \\
			& Rosenbrock  & 2.59E+07$_{\pm1.45E+07}$ & 1.37E+02$_{\pm6.25E+01}$ & 4.91E+01$_{\pm4.61E+01}$ & 9.24E+01$_{\pm1.25E+02}$ & 5.19E+01$_{\pm4.05E+01}$ & \textbf{2.89E+01$_{\pm3.56E-02}$} \\
			& Quartic     & 1.30E+01$_{\pm1.11E+01}$ & 6.08E-02$_{\pm1.58E-02}$ & 2.62E-02$_{\pm5.59E-03}$ & 1.36E-01$_{\pm4.78E-02}$ & 4.02E-02$_{\pm1.79E-02}$ & \textbf{5.21E-05$_{\pm4.97E-05}$} \\
			& Rastrigin    & 2.63E+02$_{\pm4.27E+01}$ & 8.77E+01$_{\pm7.24E+00}$ & 1.23E+02$_{\pm7.54E+01}$ & 1.13E+02$_{\pm3.74E+01}$ & 1.29E+01$_{\pm3.73E+00}$ & \textbf{1.01E-04$_{\pm8.86E-05}$} \\
			& Ackley     & 1.99E+01$_{\pm5.99E-01}$ & 2.81E+00$_{\pm3.51E-01}$ & 7.20E+00$_{\pm1.00E+01}$ & 1.79E+01$_{\pm6.89E+00}$ & 2.69E+00$_{\pm6.96E-01}$ & \textbf{1.22E-04$_{\pm9.02E-05}$} \\
			& Levy        & 1.91E+02$_{\pm6.33E+01}$ & 1.21E+00$_{\pm2.24E-02}$ & 1.10E+00$_{\pm2.02E-02}$ & 1.35E+00$_{\pm1.05E-01}$ & 1.18E+00$_{\pm7.41E-02}$ & \textbf{8.36E-05$_{\pm8.64E-05}$} \\
			\bottomrule[1pt]
			\specialrule{0em}{0.5pt}{0.5pt}
			\bottomrule[1pt]
		\end{tabular}
	}
\end{table*}
\begin{figure}[t]
	\centering
	\includegraphics[width=8cm,height=4cm, keepaspectratio]{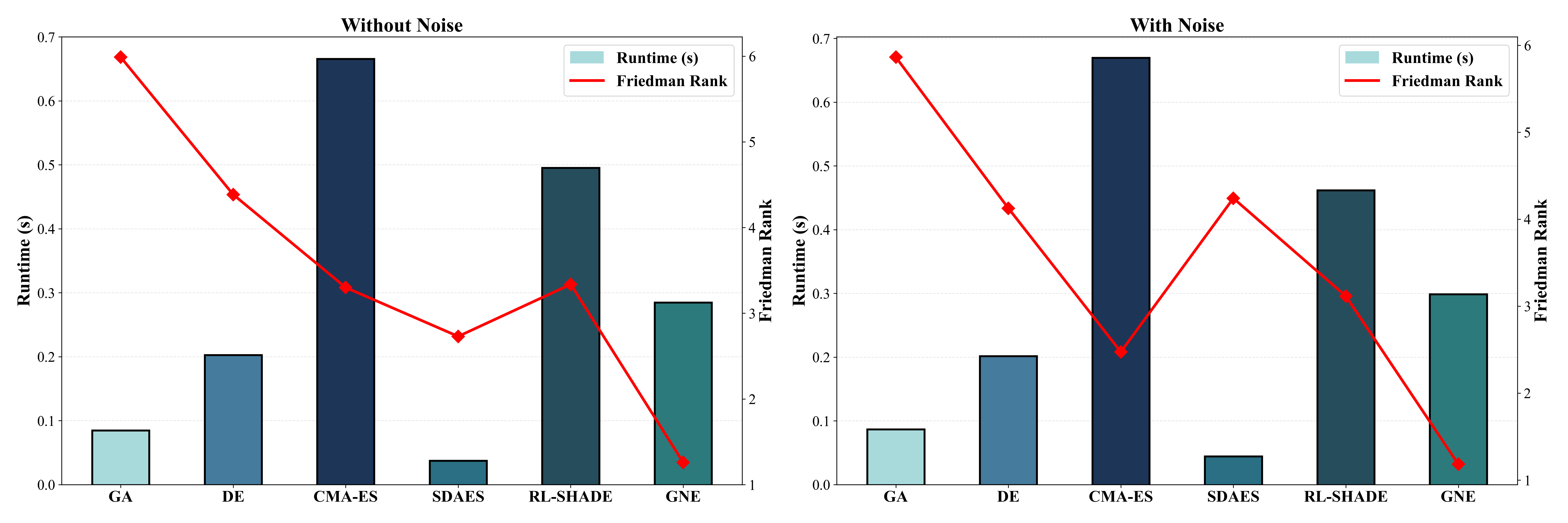}
	\vspace{-5pt}
	\caption{Comparison of the runtime and Friedman ranking of different algorithms on the benchmark function}
	\vspace{-10pt}
	\label{fig:4}
\end{figure}
\section{Experiments}
In this section, we evaluate the performance of GNE, where the filter function is represented using Chebyshev polynomials, against five well-established and widely recognized evolutionary algorithms: GA\cite{holland1992genetic},DE\cite{storn1997differential}, CMA-ES\cite{hansen2003reducing}, SDAES\cite{he2019large}, and RL-SHADE\cite{zhang2022learning}. The comparison is conducted on nine classical benchmark functions to evaluate the convergence and robustness of GNE. Detailed descriptions of these benchmark functions, including their mathematical expressions, bounds, dimensions, and characteristics.

Additionally, we examine GNE's performance under challenging scenarios, such as when noise is added to the function or when the optimal solution is shifted, to verify its robustness and adaptability in diverse optimization settings. The experiment was conducted using an Intel Core i5-12400 CPU (2.50 GHz), 16 GB of RAM, and the MATLAB 2023a platform for analysis.

\subsection{Comparison of GNE and Other Evolutionary Algorithms}
To validate the effectiveness of GNE, we compare it against classical and advanced evolutionary algorithms (GA, DE, CMA-ES, SDAES, RL-SHADE) on nine benchmark functions: Sphere, Schwefel, Schwefel 2.22, Schwefel 2.26, Rosenbrock, Quartic, Rastrigin, Ackley, and Levy. For each algorithm, the population size N is set to 30, and the maximum number of iterations T is set to 500. Each algorithm is independently executed 30 times, and we calculate the mean and standard deviation of the results.Table \ref{tab_1} presents the performance comparison of GNE with other evolutionary algorithms (GA, DE, CMA-ES, SDAES, and RL-SHADE) on various benchmark functions. Across all tested functions, GNE consistently achieves the best results in terms of both mean and standard deviation, highlighting its precision, stability, and robustness. For example, in Sphere, Schwefel, and Schwefel 2.22, GNE significantly outperforms other algorithms, achieving near-zero average values with the smallest standard deviation, demonstrating its ability to converge accuracy and stability to the global optimum. In multi-modal functions such as Rastrigin and Ackley, where the landscape is highly complex with numerous local optima, GNE excels in maintaining a balance between exploration and exploitation, delivering superior accuracy compared to other methods.On Rosenbrock and Levy, GNE outperforms all other algorithms with its ability to adapt to complex optimization spaces, achieving the lowest error and exhibiting exceptional stability. In contrast, other algorithms show varying levels of performance. While DE and CMA-ES generally perform better than GA on most functions, they often struggle to match GNE’s precision and stability. RL-SHADE and SDAES exhibit competitive performance on certain functions, such as Rosenbrock and Ackley, but their results still lack the consistency observed with GNE, particularly in terms of standard deviation, which is crucial for robustness. GA, being the simplest method, demonstrates the weakest performance overall, with higher mean and larger standard deviations across all functions, indicating its limited capacity for handling complex or multi-modal optimization problems.

\begin{table*}[t]
	\renewcommand\arraystretch{1.2}
	\caption{Performance Comparison of GNE with Other Evolutionary Algorithms on Optimal Solution Deviation Experiment}
	\label{tab_2}
	\centering
	\resizebox{0.78\textwidth}{!}{
		\begin{tabular}{cccccccc}
			\toprule[1pt]
			\specialrule{0em}{0.5pt}{0.5pt}
			\toprule[1pt]
			
			\multirow{2}{*}{\textbf{Deviation}}
			& \multicolumn{5}{c}{\textbf{Non-Graph EAs}}
			& \multicolumn{1}{c}{\textbf{Graph EAs}} \\
			\cmidrule(lr){2-6} \cmidrule(lr){7-7} & \textbf{GA} & \textbf{DE} & \textbf{CMA-ES} & \textbf{SDAES} & \textbf{RL-SHADE} & \textbf{GNE} \\
			\midrule[1pt]
			\textbf{-10} & 2.60E+04$_\text{$\pm$ 7.94E+03}$ & 3.94E-04$_\text{$\pm$ 1.63E-04}$ & 1.20E-05$_\text{$\pm$ 5.67E-06}$ & 7.77E-18$_\text{$\pm$ 9.82E-18}$ & 2.17E-07$_\text{$\pm$ 5.60E-07}$ & \textbf{4.22E-19$_\text{$\pm$ 1.33E-19}$} \\
			
			\textbf{-30} & 3.51E+04$_\text{$\pm$ 1.12E+04}$ & 5.49E-04$_\text{$\pm$ 2.49E-04}$ & 1.21E-05$_\text{$\pm$ 5.81E-06}$ & 9.00E-18$_\text{$\pm$ 1.37E-17}$ & 5.84E-08$_\text{$\pm$ 6.79E-08}$ & \textbf{4.12E-19$_\text{$\pm$ 1.12E-19}$} \\
			
			\textbf{-50} & 5.47E+04$_\text{$\pm$ 1.19E+04}$ & 4.01E-04$_\text{$\pm$ 1.53E-04}$ & 1.41E-05$_\text{$\pm$ 8.43E-06}$ & 5.05E-18$_\text{$\pm$ 6.04E-18}$ & 4.55E-07$_\text{$\pm$ 1.79E-06}$ & \textbf{3.86E-19$_\text{$\pm$ 1.17E-19}$} \\
			
			\textbf{-70} & 1.04E+05$_\text{$\pm$ 2.16E+04}$ & 3.49E-04$_\text{$\pm$ 1.49E-04}$ & 1.63E-05$_\text{$\pm$ 8.97E-06}$ & 3.79E-17$_\text{$\pm$ 9.31E-17}$ & 2.03E-07$_\text{$\pm$ 3.91E-07}$ & \textbf{4.39E-19$_\text{$\pm$ 1.26E-19}$} \\
			
			\textbf{-90} & 1.71E+05$_\text{$\pm$ 2.51E+04}$ & 1.31E-04$_\text{$\pm$ 7.58E-05}$ & 1.92E-05$_\text{$\pm$ 8.10E-06}$ & 5.65E-17$_\text{$\pm$ 1.38E-16}$ & 1.15E-07$_\text{$\pm$ 2.22E-07}$ & \textbf{4.06E-19$_\text{$\pm$ 1.00E-19}$} \\
			
			\textbf{+10} & 2.22E+04$_\text{$\pm$ 8.65E+03}$ & 3.97E-04$_\text{$\pm$ 2.13E-04}$ & 1.31E-05$_\text{$\pm$ 6.11E-06}$ & 9.38E-18$_\text{$\pm$ 1.45E-17}$ & 2.16E-07$_\text{$\pm$ 5.23E-07}$ & \textbf{4.09E-19$_\text{$\pm$ 1.01E-19}$} \\
			
			\textbf{+30} & 2.10E+04$_\text{$\pm$ 8.80E+03}$ & 4.88E-04$_\text{$\pm$ 1.60E-04}$ & 1.13E-05$_\text{$\pm$ 4.82E-06}$ & 9.81E-18$_\text{$\pm$ 1.94E-17}$ & 2.71E-07$_\text{$\pm$ 4.49E-07}$ & \textbf{4.30E-19$_\text{$\pm$ 9.91E-20}$} \\
			
			\textbf{+50} & 2.60E+04$_\text{$\pm$ 1.22E+04}$ & 4.60E-04$_\text{$\pm$ 1.49E-04}$ & 1.59E-05$_\text{$\pm$ 6.11E-06}$ & 7.66E-18$_\text{$\pm$ 9.10E-18}$ & 6.21E-07$_\text{$\pm$ 2.51E-06}$ & \textbf{4.20E-19$_\text{$\pm$ 1.20E-19}$} \\
			
			\textbf{+70} & 4.76E+04$_\text{$\pm$ 1.60E+04}$ & 3.36E-04$_\text{$\pm$ 1.26E-04}$ & 1.60E-05$_\text{$\pm$ 7.53E-06}$ & 1.42E-17$_\text{$\pm$ 2.63E-17}$ & 1.09E-07$_\text{$\pm$ 1.19E-07}$ & \textbf{3.79E-19$_\text{$\pm$ 6.79E-20}$} \\
			\textbf{+90} & 1.10E+05$_\text{$\pm$ 2.67E+04}$ & 1.42E-04$_\text{$\pm$ 5.28E-05}$ & 1.99E-05$_\text{$\pm$ 7.86E-06}$ & 3.19E-17$_\text{$\pm$ 7.22E-17}$ & 8.24E-08$_\text{$\pm$ 2.55E-07}$ & \textbf{3.79E-19$_\text{$\pm$ 1.55E-19}$} \\
			
			\bottomrule[1pt]
			\specialrule{0em}{0.5pt}{0.5pt}
			\bottomrule[1pt]
		\end{tabular}
	}
\end{table*}

\subsection{Robustness of GNE in Noisy Function Environments}

In real-world scenarios, function evaluations are often affected by measurement errors and uncertainties\cite{qian2018effectiveness,qian2018analyzing}. In this section, we introduce random noise uniformly distributed in the range [0, 1] to the nine benchmark functions to evaluate the performance of GNE under noisy function conditions. This setup tests the algorithm’s robustness when noise is present in the computation of function values.

Table \ref{tab_1} highlights the performance of GNE and other evolutionary algorithms (GA, DE, CMA-ES, SDAES, and RL-SHADE) under noisy benchmark functions, showcasing their robustness in handling noisy environments. Compared to the noise-free scenarios in Table \ref{tab_1}, the introduction of random noise significantly impacts the performance of most algorithms.GA and SDAES experience severe performance degradation, with large mean and high standard deviations across almost all functions, indicating their inability to effectively handle noise. DE and CMA-ES also show reduced performance, particularly in complex multi-modal functions like Rastrigin-Noise and Ackley-Noise, where noise amplifies the difficulty of optimization.  RL-SHADE performs relatively better than most algorithms, but its increased variability under noisy conditions makes its results less reliable. In stark contrast, GNE consistently achieves the smallest averages and standard deviations across all noisy benchmark functions, maintaining its superior performance even in the presence of noise. For example, in functions such as Sphere-Noise, Schwefel-Noise, and Quartic-Noise, GNE significantly outperforms other algorithms, demonstrating its robustness and ability to effectively mitigate the influence of noise.  Even in challenging scenarios like Rastrigin-Noise and Ackley-Noise, where noise typically causes significant performance degradation, GNE remains highly accurate and stable. This exceptional robustness can be attributed to GNE's capture of global correlations within the population, controlling global-scale features through frequency components. It avoids algorithm failure caused by excessive fitness evaluation noise of a few individuals or improper parameter selection, thereby enhancing the algorithm's adaptability to external environmental uncertainties.

To demonstrate the balance between overall performance and computational time in GNE, Figure \ref{fig:4} presents the Friedman ranking\cite{friedman1937use} and average running times of different algorithms on benchmark functions under both noise-free and noisy conditions. GA and DE, two classic and simple algorithms, exhibit the shortest running times in both conditions but achieve the lowest rankings. CMA-ES and RL-SHADE rank in the middle under both conditions but require the longest and second-longest running times, respectively.SDAES achieves the shortest running time in the noise-free condition and ranks second, slightly behind GNE. However, in the presence of noise, its optimization efficiency significantly deteriorates, dropping to the second-to-last position, even performing worse than the simple algorithm DE. In contrast, GNE achieves the highest ranking under both noise-free and noisy conditions while using a moderate amount of computational time. Additionally, GNE demonstrates strong adaptability to the optimization of functions with added noise, maintaining robust performance across varying conditions.

\subsection{Robustness of GNE for Optimal Solution Deviation}

A critical problem in benchmarking and analysis of evolutionary computation methods is that many benchmark functions have optima located at the center of the feasible set, which can create biases in the evaluation of evolutionary algorithms\cite{kudela2022critical}.The center-bias in some methods allows them to easily find optima, rendering comparisons with other algorithms that lack such biases ineffective. To address this issue, we study whether the proposed GNE algorithm exhibits overfitting to the optimal solution by testing its performance under varying deviations. The optimal solution is intentionally shifted by deviations of -10, -30, -50, -70, -90, +10, +30, +50, +70, and +90. This setup is designed to test GNE's robustness when the optimal solution deviates from its original position.Table \ref{tab_2} presents the performance comparison of GNE with other evolutionary algorithms (GA, DE, CMA-ES, SDAES, and RL-SHADE) under different optimal solution deviations in the Sphere function. This highlights the robustness and adaptability of GNE in handling dynamic changes in the solution landscape, thanks to the translation invariance property of its adjacency matrix, which captures global correlations by modeling relative relationships within the population without relying on the boundaries of the solution space.

\section{Conclusion}
In this study, we present the GNE framework, which models pairwise relationships within the population by means of an adjacency matrix and captures global correlations via spectral decomposition.   Evolutionary signals are encoded into frequency components whose global-scale influence is modulated by a polynomial filter function, thereby rendering the exploration–exploitation balance both interpretable and effective. We demonstrate GNE's performance on nine benchmark functions (e.g., Sphere, Rastrigin, and Rosenbrock), where it consistently outperforms classical algorithms (GA, DE, CMA-ES) and advanced algorithms (SDAES, RL-SHADE) under various conditions, including noise corruption and deviations from the optimal solution.   By synthesizing the experimental results, we demonstrate the effectiveness of GNE within existing evolutionary methods and argue that evolution via spectral graph representations enables explicit modeling of evolutionary processes. Future work will focus on testing GNE’s performance on larger-scale optimization problems, exploring ways to reduce the complexity of spectral decomposition (e.g., adaptive population size reduction), and applying GNE to more real-world optimization problems, such as complex engineering problems and optimization of neural network parameters and structures.

\section{Acknowledgments}
This work was supported by the Fundamental Research Funds for the Central Universities (No. N25XQD062) and Open Research Projects of the Key Laboratory of Blockchain Technology and Data Security of the Ministry of Industry and Information Technology for the year 2025 (No. KT20250015). Corresponding authors: Puning Zhao and Dayu Hu.

\bibliography{GNECITE}

\end{document}